\documentclass[preprint,12pt]{elsarticle}

\usepackage{amssymb}
\usepackage{amsmath}

\usepackage{amsmath,amsfonts}
\usepackage{algorithmic}
\usepackage{algorithm}
\usepackage{array}
\usepackage[caption=false,font=normalsize,labelfont=sf,textfont=sf]{subfig}
\usepackage{textcomp}
\usepackage{stfloats}
\usepackage{url}
\usepackage{verbatim}
\usepackage{graphicx}
\usepackage{tabularx}

\usepackage{siunitx} 
\usepackage{booktabs}
\usepackage[hidelinks]{hyperref}

\begin{document}

\begin{frontmatter}



\title{Fully Differentiable Lagrangian Convolutional Neural Network for Physics-Informed Precipitation Nowcasting}


\author[kinit,vut]{Peter Pavlík\texorpdfstring{\corref{pp}}}
\ead{peter.pavlik@kinit.sk}
\author[kinit]{Martin Výboh}
\ead{martin.vyboh@kinit.sk}
\author[kinit]{Anna Bou Ezzeddine}
\ead{anna.bou.ezzeddine@kinit.sk}
\author[kinit]{Viera Rozinajová}
\ead{viera.rozinajova@kinit.sk}

\cortext[pp]{Corresponding author}

\affiliation[kinit]{organization={Kempelen Institute of Intelligent Technologies},
            addressline={Bottova~7939/2A}, 
            city={Bratislava},
            postcode={811~09},
            country={Slovakia}}
\affiliation[vut]{organization={Faculty of Information Technology, Brno University of Technology},
            addressline={Božetěchova~2}, 
            city={Brno},
            postcode={612~00},
            country={Czech Republic}}

\begin{abstract}
This paper presents a convolutional neural network model for precipitation nowcasting that combines data-driven learning with physics-informed domain knowledge. We propose LUPIN, a Lagrangian Double U-Net for Physics-Informed Nowcasting, that draws from existing extrapolation-based nowcasting methods. It consists of a U-Net that dynamically produces mesoscale advection motion fields, a differentiable semi-Lagrangian extrapolation operator, and an advection-free U-Net capturing the growth and decay of precipitation over time. Using our approach, we successfully implement the Lagrangian convolutional neural network for precipitation nowcasting in a fully differentiable and GPU-accelerated manner. This allows for end-to-end training and inference, including the data-driven Lagrangian coordinate system transformation of the data at runtime. We evaluate the model and compare it with other related AI-based models both quantitatively and qualitatively in an extreme event case study. Based on our evaluation, LUPIN matches and even exceeds the performance of the chosen benchmarks, opening the door for other Lagrangian machine learning models.
\end{abstract}



\begin{keyword}
Rain \sep Forecasting \sep Neural networks \sep Meteorological radar


\end{keyword}

\end{frontmatter}



\newtheorem{theorem}{Theorem}

\section{Introduction} \label{sec:intro}

Precipitation nowcasting is the task of predicting the future intensity of precipitation (rainfall, snowfall, hail, etc.) in the near future with high local detail~\cite{WMO17}. The nowcasts are valuable for early warning about extreme events and can help mitigate damage and loss of lives. Accurate precipitation nowcasting can therefore increase our resilience to the climate crisis, particularly in the face of increased precipitation variability~\cite{Zhang2024}. Radar precipitation observations from ground radar stations are good approximations of the current rainfall rate and therefore are usually used as a basis for precipitation nowcasting. 

Traditionally, the nowcasting is performed using radar extrapolation methods based on the assumption that precipitation at the mesoscale behaves and advects according to the continuity equation. A few past precipitation observations are used to estimate an advection motion field and the last observation is extrapolated forward in time according to this field. This is the basis of many operational methods.

However, in the last few years, convolutional neural networks (CNNs) have become the de-facto state-of-the-art models for precipitation nowcasting. The first application of a CNN-based architecture in the domain was the \mbox{ConvLSTM} model~\cite{shi2015convolutional}. It leveraged that the problem can be posed as a spatiotemporal sequence forecasting problem similar to next-frame video prediction and therefore is a good fit for processing using CNNs.

One obvious problem with this architecture is that the convolutional recurrence structure in ConvLSTM is location-invariant, therefore not explicitly taking into consideration the advection motion of precipitation over time. This led to the development of the Trajectory GRU (TrajGRU) model~\cite{shi2017deep} that can learn the location-variant structure for recurrent connections, therefore internally considering the precipitation advection. It was shown that the TrajGRU is more efficient in capturing the spatiotemporal correlations in the data.

Despite the TrajGRU model seemingly combining the best of both worlds -- the physics-informed location-variant structure with data-driven CNN-based processing -- location-invariant architectures continued to be developed and refined.

In \cite{ayzel2020rainnet}, a U-Net architecture was successfully applied for precipitation nowcasting to create RainNet. Despite being a purely data-driven model without explicit inclusion of domain knowledge, it performs well and is often viewed as a simple benchmark CNN-based model.

The main challenge for all mentioned data-driven nowcasting models is the growing smoothness and fading out of the nowcasts quantities with growing time horizons. Due to the highly chaotic behavior of the precipitation in the atmosphere, there is a high degree of uncertainty in the predictions. When training with traditional gridpoint-based loss functions, we face the so-called “double penalty problem”. A forecast of a precipitation feature that is correct in terms of intensity, size, and timing, but incorrect concerning location, results in a very large error~\cite{Keil2009}. This leads to "mathematically optimal" predictions being visibly blurry and physically inconsistent, limiting their usefulness as nowcasts for early warning of extreme events, as those especially are smoothed out.

In an attempt to solve the blurriness challenge, some models introduced a GAN-based framework to produce realistically looking nowcasts~\cite{tian2019generative, wang2021using}. A notable example is the Deep Generative Model of Radar (DGMR) model~\cite{ravuri2021skilful} that is able to produce accurate and realistic nowcasts up to 90 minutes ahead. A visual verification by meteorologists was used, and it was selected as the best in most cases by a large margin out of all the compared nowcasting models at the time.

Recently, the GAN approach was combined with the explicit consideration of advection motion fields in NowcastNet~\cite{zhang2023skilful}. It incorporates a differentiable evolution network that generates a motion field and performs a deterministic evolution of the precipitation according to the continuity equation before passing it to the main generative model. This helped the model outperform the DGMR by a large margin in expert evaluation at up to 3-hour ahead lead times.

Another recent model that explicitly takes into consideration the advection of the precipitation is the Lagrangian CNN (L-CNN) introduced in~\cite{lcnn}. Here, the advection motion field obtained from an optical flow algorithm is used to transform the input into the Lagrangian coordinate system, where a U-Net processes the temporally differenced advection-free inputs, thus separating the learning of rainfall growth and decay from mesoscale motion. The model outperformed a benchmark U-Net-based RainNet, showing that incorporating insights from rainfall physics can lead to substantial improvements in machine learning nowcasting methods. However, due to the motion field being a product of a non-differentiable optical flow method, the model can not be trained using a fully end-to-end approach, constraining its utility.

In this paper, we present a new model that takes inspiration from the differentiable evolution network of NowcastNet and applies it to the ingenious Lagrangian advection-free approach of L-CNN to form a single end-to-end CNN-based model. We implement a differentiable semi-Langrangian extrapolation that allows us to train the model to produce an advection motion field and dynamically map the inputs into the Lagrangian coordinates to provide a better insight for the subsequent learning of precipitation evolution by performing the Lagrangian coordinate transformation and temporal differencing during runtime. We present LUPIN: A \textbf{L}agrangian Double \textbf{U}-Net for \textbf{P}hysics-\textbf{I}nformed \textbf{N}owcasting.

\section{The Lagrangian View}

To understand the theoretical underpinnings of our approach, we first present the assumptions and methods on which the traditional nowcasting methods are based. The formalism is mostly based on the notations of~\cite{germann2002scale} and~\cite{lcnn}.

\subsection{Advection Equation and Lagrangian Persistence}

The precipitation in the near future is highly dependent on the present weather. The basis of all traditional nowcasting methods is therefore the assumption of persistence -- the precipitation in the near future will stay roughly the same as observed in the present.

In the Eulerian coordinates (where we look at fluid flow in relation to the space that it moves through) this means the simplest produced nowcast is an observation identical to the last one. However, the performance of this Eulerian persistence "model" falls rather quickly with increasing lead times as the current precipitation moves elsewhere.

The assumption of persistence is more realistic under the Lagrangian specification of the flow field -- where we follow fluid parcels as they move through space and time. Lagrangian persistence assumes the overall shape of the precipitation will stay the same but accounts for the displacement of the precipitation according to the advection equation. The advection equation can be expressed mathematically as follows:

\begin{equation} \frac{\partial \Psi }{\partial t}+\nabla \cdot (\Psi \boldsymbol{u})=0 \label{eq:adv_eq} \end{equation}

where $\Psi$ denotes the 2D radar precipitation field and $\boldsymbol{u}$ the horizontal motion field. It is often assumed that the flow is incompressible, therefore the divergence of the motion field is zero ($\boldsymbol{u}$ is solenoidal):

\begin{equation} \nabla \boldsymbol{u}=0 \label{eq:divergence}\end{equation}

Under this assumption, the advection equation (\ref{eq:adv_eq}) can be written in the form:

\begin{equation} \frac{\partial \Psi }{\partial t}+\boldsymbol{u}\cdot \nabla \Psi=0\end{equation}

We can transform the precipitation field $\Psi$ to a Lagrangian coordinate system precipitation field $\widetilde{\Psi }$ by extrapolating it according to the motion field $\boldsymbol{u}$ so that $\boldsymbol{u}\cdot \nabla\widetilde{\Psi}=0$ holds and the advection equation can finally be written as:

\begin{equation} \frac{\partial\widetilde{\Psi}}{\partial t}=0\label{eq:adv_eq_lagrange}\end{equation}

The advection equation now clearly denotes the Lagrangian persistence assumption, expressing that the precipitation field in Lagrangian coordinates will not change over time.

\subsection{Implementation of the Lagrangian Transformation}

To transform the input 2D precipitation fields $\Psi_{1,2, ...,n}$ to the Lagrangian coordinates $\widetilde{\Psi}_{1,2, ...,n}$, all the inputs need to be extrapolated to a single reference field. We define an extrapolation operator $\xi^t(\Psi, \boldsymbol{u})$ as an operation of extrapolating a precipitation field $\Psi$ by $t$ time steps forward in time along the motion field $\boldsymbol{u}$.

Semi-Lagrangian extrapolation~\cite{sawyer1963semi} is commonly used as an implementation of this operator. For estimating the motion field $\boldsymbol{u}$, various optical flow methods are available, the Lucas-Kanade algorithm~\cite{lucas1981iterative, bouguet2001pyramidal} being a popular one. Both of these are implemented in Python language in the pySTEPS nowcasting library~\cite{gmd-12-4185-2019}.

Formally, to transform $n$ successive input precipitation fields to Lagrangian coordinates (choosing the last one as reference), the extrapolation operator must be applied to each input image the number of times equal to the time step difference between it and the reference image:

\begin{equation}\widetilde{\Psi}_{i}=\xi^{n-i}(\Psi_{i}, \boldsymbol{u}),\quad i=1,2, \ldots, n-1 \label{eq:extra}
\end{equation}

After this transformation, we can estimate the actual temporal difference of precipitation field from the equation (\ref{eq:adv_eq_lagrange}) at a given time $t$ as:

\begin{equation}\frac{\partial \widetilde{\Psi }}{\partial t} \approx \Delta \widetilde{\Psi }_{t} = \widetilde{\Psi }_{t} - \widetilde{\Psi }_{t-1}
\end{equation}

\subsection{The Lagrangian Nowcasting}

For a more realistic nowcasting of the rapidly-changing evolution of precipitation over time, we need to introduce one more term to the advection equation -- a source-sink term $S_{\Psi}$ which allows for the refinement of the Lagrangian persistence assumption:

\begin{equation} \frac{\partial\widetilde{\Psi}}{\partial t}=S_{\Psi}\label{eq:source-sink}\end{equation}

It represents the changes in the radar precipitation field over time without considering its displacement according to the advection motion field. It can therefore be considered the the advection-free change of the field over time.

However, predicting the source-sink term $S_{\Psi}$ and therefore the growth and decay of precipitation systems is very difficult. It is the biggest challenge of the nowcasting systems.

The state-of-the-art extrapolation nowcasting methods such as S-PROG~\cite{seed2003dynamic}, LINDA~\cite{pulkkinen2021lagrangian}, STEPS~\cite{bowler2006steps} and ANVIL~\cite{pulkkinen2020nowcasting} include an autoregressive model to estimate the source-sink term $S_{\Psi}$ based either on the input precipitation fields in Lagrangian coordinates $\widetilde{\Psi}$ or the estimated time derivatives $\Delta \widetilde{\Psi}$.

\subsection{Incorporating Convolutional Neural Networks}

Analogously to the autoregressive approaches,  some CNN-based models have taken the Lagrangian approach to estimate source-sink term $S_{\Psi}$.

In the case of TrajGRU~\cite{shi2017deep}, the model implicitly learns the flow field $\boldsymbol{u}$ via the dynamically determined neural connections, therefore learning to perform the right extrapolation before estimating the source-sink term.

The L-CNN model~\cite{lcnn} takes a different approach, transforming the precipitation fields to Lagrangian coordinates before the training of the model (based on Lucas-Kanade advection motion field) and time differencing the transformed fields. Then they train a U-Net to predict the future temporal differences $\Delta \widetilde{\Psi}$/$S_{\Psi}$.

Another model worth mentioning here is the NowcastNet~\cite{zhang2023skilful}, specifically its Evolution Network. While it does not transform the precipitation maps to Lagrangian coordinates, it learns to separate the prediction of advection motion field from the prediction of changes in precipitation field intensity, analogous to the source-sink term $S_{\Psi}$. Also, it introduces a motion-regularization term loosely based on the continuity (advection) equation (\ref{eq:adv_eq}) to enforce a smoother motion field. In contrast to L-CNN, the differentiable evolution operator allows this model to be trainable in an end-to-end manner.

\begin{figure}
    \centering
    \includegraphics[width=0.45\columnwidth]{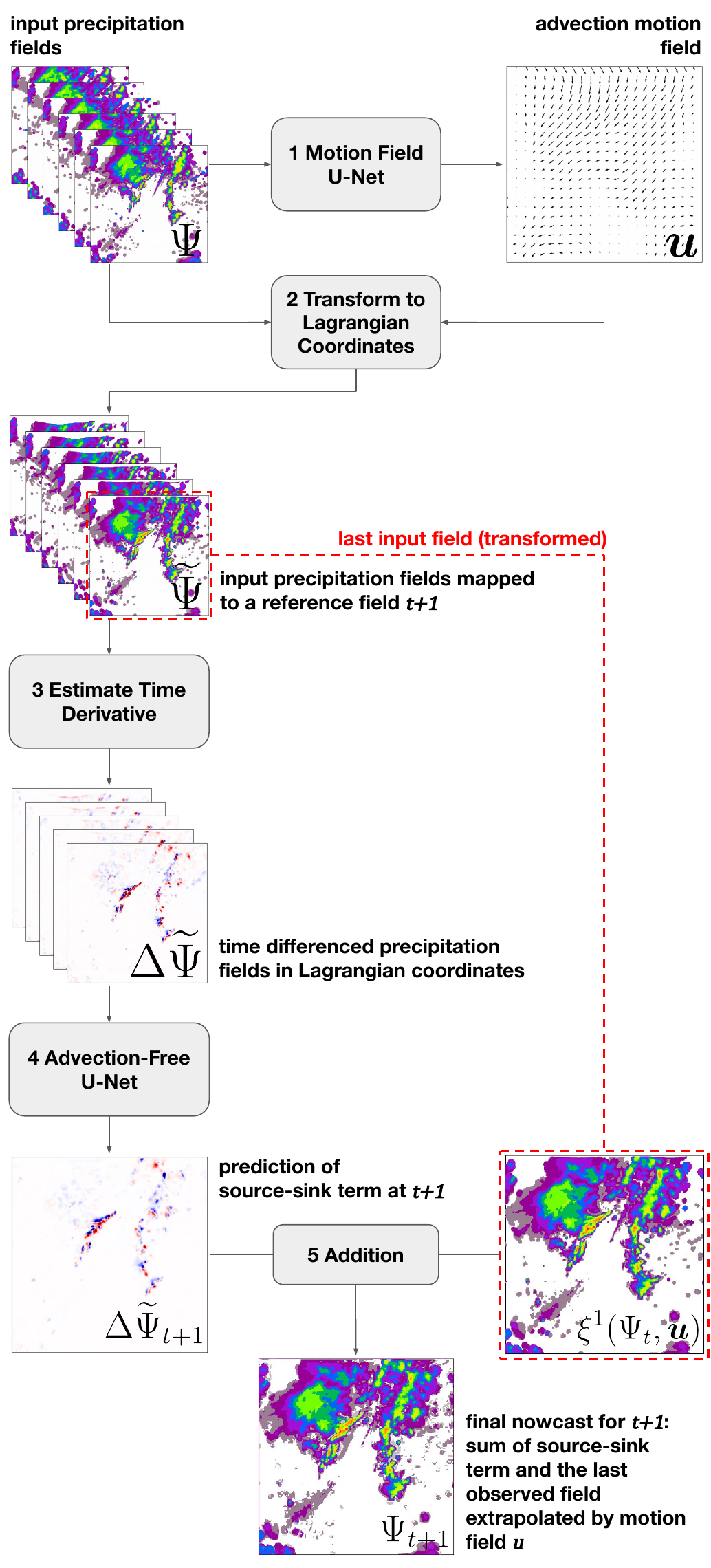}
    \caption{Architecture of the LUPIN model and the inference process of a nowcast one time step ahead. The gray processing blocks are marked 1 to 5 in the order they are called during inference. \textbf{1}~The first step is the calculation of the advection motion field $\boldsymbol{u}$. \textbf{2}~Then, the $n$ input fields $\Psi$ can be transformed into Lagrangian coordinates as described in Equation \ref{eq:extra}. \textbf{3}~The precipitation maps in Lagrangian coordinates are differenced over time to obtain $n-1$ time differenced fields $\Delta \widetilde{\Psi}$. \textbf{4}~The differenced fields are used as an input to the Advection-Free U-Net that predicts the next differenced field $\Delta \widetilde{\Psi}_{t+1}$ (representing the source-sink term $S_{\Psi}$ at time $t+1$). \textbf{5}~The predicted source-sink term is added to the last input field (already transformed in step 2 to reference field $t+1$), the product being the final nowcast for the next time step.}
    \label{fig:arch}
\end{figure}

\section{LUPIN -- Lagrangian Double U-Net for Physics-Informed Nowcasting}

This section is dedicated to introducing and describing the architecture and training process of our precipitation nowcasting model LUPIN. We compare it to previous approaches, explain the differences, and elaborate on the thought process behind them. The main point of comparison will be the Lagrangian CNN (L-CNN) model from \cite{lcnn}, as it served as the main inspiration for our approach, whose shortcomings we aim to address.

\subsection{LUPIN Architecture}

The overall architecture of the LUPIN model is based on the architecture of the L-CNN model~\cite{lcnn}. We took the key idea of the L-CNN -- to transform the input data to Lagrangian coordinates and use a CNN to model the time derivatives of rainfall fields (the source-sink term $S_{\Psi}=\Delta \widetilde{\Psi}_{t+1}$) -- and decided to explore implementing this approach as a fully data-driven model.

The inclusion of the optical flow algorithm in the L-CNN model makes the training process cumbersome. The algorithm is orders of magnitude slower compared to the time needed for GPU-accelerated inference and backpropagation in a CNN. This makes the training feasible only with inputs and outputs transformed to Lagrangian coordinates beforehand. At the same time, the longer inference limits the application of the model for ensemble nowcasting.

Our solution was to implement a differentiable semi-Lagrangian-like extrapolation operation in the PyTorch machine learning framework~\cite{paszke2019pytorch}. This allowed us to create a model that learns to produce the motion field in a data-driven manner and then performs the transformation to Lagrangian coordinates dynamically during runtime in a fraction of the time.

The overall LUPIN architecture is in the form of a double U-Net, see Figure \ref{fig:arch}. The first U-Net (Motion Field U-Net) produces the motion fields and functionally replaces the Lukas-Kanade optical flow algorithm in the L-CNN. The second one (Advection-free U-Net) works with the Lagrangian-transformed and time differenced inputs analogous to the L-CNN specification.

The model is to be applied iteratively in a rolling window manner. Each inference produces a new motion field and the corresponding nowcast for the next time step. The nowcast then subsequently serves as the last input observation for the next iteration. By choosing the reference time for the Lagrangian mapping to match the one-step ahead nowcast window $t+1$, we have eliminated the need to perform the transformation back to Eulerian coordinates as the final prediction step, simplifying the inference process in comparison to the L-CNN.

\subsection{Training Process}

Training the LUPIN is a three-stage process. While it is feasible to train the whole model at once, preliminary experiments have shown this to be unstable. We therefore used the following process:

\begin{enumerate}
  \item Train the Motion Field U-Net (MF-U-Net) to produce satisfactory motion fields by itself.
  \item Train the Advection-Free U-Net (AF-U-Net) with the MF-U-Net's weights frozen.
  \item Unfreeze MF-U-Net's weights and continue the training of both U-Nets together to act jointly as a single model.
\end{enumerate}

\subsubsection{Training MF-U-Net}

Training the MF-U-Net to produce physically consistent advection motion fields $\mathbf{u}$ is not trivial. To be usable in the Lagrangian framework, the motion fields should be stable over time when produced iteratively and locally smooth so as to not interfere in the subsequent estimation of temporal differences in Lagrangian coordinates. We want to emulate a simple Lagrangian persistence model in a data-driven way.

The first intuitive approach was to just minimize the error between the last input observation extrapolated by one time step and the ground truth. We denote the error criterion $\mathcal{C}$ and the target precipitation field $\hat{\Psi}$.

\begin{equation} \mathcal{L}_{naive}=\mathcal{C}(\xi^{1}(\Psi_{t}, \boldsymbol{u}), \hat{\Psi}_{t+1})\end{equation}

However, regardless of the criterion $\mathcal{C}$, this loss function produces highly unrealistic motion fields full of sinks and sources, functionally interfering with the future task of the AF-U-Net. It goes against the assumption described by Equation~\ref{eq:divergence} of the divergence-free motion field for the Lagrangian persistence.

To mitigate this, we introduce two modifications to the training process. Firstly, we produce a pooled optimal motion field by minimizing the error of extrapolating the precipitation fields by one-time step over the whole input sequence, not just the last observation and the one following it.

\begin{equation} \mathcal{L}_{MF}=\sum_{t=1}^{n}\mathcal{C}(\xi^{1}(\Psi_{t}, \boldsymbol{u}), \hat{\Psi}_{t+1})\end{equation}

The loss function above iterates over all $n$ input observations and sums the error of extrapolating them one step forward in time. This pushes the model to produce more stable motion fields better fit for transforming the inputs into the Lagrangian coordinate system.

Secondly, we introduce a physics-informed regularization loss to directly penalize breaking the Equation \ref{eq:divergence}. By enforcing it, we can push the model towards creating solutions consistent with the continuity equation derived from conservation laws. The assumption is also one of the Navier-Stokes partial differential equations that govern fluid dynamics \cite{batchelor1967introduction}. Since our data captures the movement of air in the atmosphere, it should behave according to these equations. The regularization will therefore make the motion fields more physically consistent.

The divergence $\nabla\boldsymbol{u}$ is calculated using partial derivatives of the motion field:

\begin{equation} \nabla\boldsymbol{u}=\frac{\partial u_x}{\partial x} + \frac{\partial u_y}{\partial y}\end{equation}

We can compute discrete approximations of motion field partial derivatives using 2D convolutions with a $3\times3$ Sobel filter~\cite{kanopoulos1988design}.

\begin{center}
a)\quad
\begin{tabular}{|c|c|c|}
 \hline
  -1 & 0 & 1 \\ 
  \hline
  -2 & 0 & 2 \\ 
  \hline
  -1 & 0 & 1 \\ 
 \hline
\end{tabular}
\qquad
b)\quad
\begin{tabular}{|c|c|c|}
 \hline
  -1 & 2 & -1 \\ 
  \hline
  0 & 0 & 0 \\ 
  \hline
  1 & 2 & 1 \\ 
 \hline
\end{tabular}
\end{center}

The filter a) is used to estimate $\frac{\partial u_x}{\partial x}$ and the filter b) estimates $\frac{\partial u_y}{\partial y}$. Using these values, we define the physics-informed continuity-consistency loss function at every point of the motion field as:

\begin{equation} \mathcal{L}_{PI}=|\nabla \boldsymbol{u}|\label{eq:reg}\end{equation}

In the vocabulary of physics-informed machine learning, this is a soft constraint or learning bias, meaning we allow the network to break the constraint, but penalize it as it diverges~\cite{karniadakis2021physics}.

The final loss function that is minimized is simply the weighted sum of the data-driven and physics-informed losses:

\begin{equation} \mathcal{L}=(1-\beta)\mathcal{L}_{MF}+\beta\mathcal{L}_{PI} \quad \beta \in (0,1)
\label{eq:loss_MF}
\end{equation}

\subsubsection{Training AF-U-Net}

The second training stage starts with taking the pre-trained MF-U-Net and freezing the weights, so as not to unlearn the motion field generation from unstable gradients being backpropagated from the downstream AF-U-Net. Because the MF-U-Net has been trained to effectively perform the Lagrangian persistence nowcasting, we can train the AF-U-Net to post-process its predictions by learning to produce a source-sink term $S_{\Psi}$ capturing the growth and decay of precipitation fields over time.

As shown in the L-CNN paper, the Lagrangian-transformed and time-differenced fields can allow the Advection-Free U-Net to better grasp the precipitation growth and decay in the data, resulting in more accurate predictions with less blurring as lead time increases. Therefore, the input fields to our Advection-Free U-Net are also being transformed and time-differenced. The Lagrangian transformation is performed dynamically during training using our implementation of the semi-Lagrangian extrapolation.

The prediction task the Advection-Free U-Net performs is in essence auto-regressive, as the target output (source-sink term $S_{\Psi t+1} = \Delta \widetilde{\Psi }_{t+1}$) is the expected difference of the last input extrapolated by one time step ($\xi^{1}(\Psi_{t}, \boldsymbol{u})$) from the ground truth observation ($\hat{\Psi}_{t+1}$). To obtain the nowcast for the next time step, a simple addition of these fields is performed:

\begin{equation}\Psi_{t+1}=S_{\Psi t+1}+\xi^{1}(\Psi_{t}, \boldsymbol{u})\end{equation}

The error of this nowcast in relation to the observed precipitation field target forms the loss function during this training stage:

\begin{equation} \mathcal{L}_{AF}=\mathcal{C}(\Psi_{t+1}, \hat{\Psi}_{t+1})\end{equation}

In a way, this can be viewed as a form of residual learning, with the output of the AF-U-Net serving as a residual being added to a single skip-connected input channel.

\subsubsection{Training LUPIN}

By training the models only separately, an issue of out-of-distribution samples arises. The AF-U-Net produces nowcasts that get blurrier as lead time increases due to uncertainty. This means that, eventually, the outputs do not look like the input images from the train set. This poses a problem for the MF-U-Net in later iterations when it is given input data that does not look like the data it was trained on, resulting in unstable advection motion fields.

Therefore, for these models to cooperate together, a final stage of training is needed. This fine-tuning stage is mostly similar to the previous one. The only change is that we unfreeze the weights of the MF-U-Net and add its loss functions to the total loss function being minimized.

In this training stage, the MF-U-Net parameters are updated not only by the gradients backpropagated from the loss function described in Equation \ref{eq:loss_MF}, but also the gradients from the AF-U-Net due to the differentiable implementation of the whole LUPIN model. We hypothesize that this might allow the model to produce more useful advection motion fields than the general optical flow algorithms, as the model is trained specifically to produce motion fields for the optimal Lagrangian coordinate transformation of the precipitation fields.

The total weighted loss optimized in this stage, consisting of the three separate weighted loss terms, is specified as:

\begin{equation} \mathcal{L}=(1-\beta)((1-\gamma)\mathcal{L}_{AF}+\gamma\mathcal{L}_{MF})+\beta\mathcal{L}_{PI}\quad \beta,\gamma \in (0,1)
\label{eq:fina_loss}
\end{equation}

\section{The Precipitation Dataset} \label{sec:dataset}
For training the LUPIN model, we use a dataset provided by the Slovak Hydrometeorological Institute (SHMU). The dataset consists of radar reflectivity observations from the 1st of January 2016 until the 30th of June 2019, collected periodically in 5-minute intervals. Data originates from Malý Javorník weather radar station, located in the Carpathian Mountains in Western Slovakia. The dataset contains $355 761$ unique observations in ODIM HDF5 format. 

The radar station captures precipitation by sending out radar beams and measuring the echo returned after bouncing off the precipitation particles in the atmosphere. This helps to determine the volume or type of precipitation present in the air. Intensity of the returned echo is called radar reflectivity ($Z$). It is measured in logarithmic dimensionless units called decibels related to $Z$ ($\si{dBZ}$). Dataset consists of reflectivity observations at multiple data points captured by radar called gates or bins. Reflectivity gates are measured at multiple horizontal and vertical angles relative to the station, resulting in reflectivity data encoded in polar coordinates.

The data is processed from raw radar reflectivity observations to Cartesian coordinates using the Py-ART Python library \cite{pyart}. We aggregate the 3-dimensional precipitation observations into a single two-dimensional reflectivity map by a method called CMAX. It groups data by returning the maximum value of reflectivity in the vertical column. Resulting reflectivity map is a $340 \si{km} \times 340 \si{km}$ sized center slice with $1 \si{km} \times 1 \si{km}$ resolution. The reflectivity is converted to rainfall rate, measured in mm/h, using Marshall-Palmer formula \cite{marshall-palmer}:
\begin{theorem}
    \label{the:m-p}
    For the reflectivity factor $Z$ and rainfall rate $R$ in $\si{mm.h^{-1}}$ holds the equation $Z = 200R^{1.6}$, where $200$ and $1.6$ are empirically derived constants.
\end{theorem}

\subsection{Dataset filtering} \label{sub:dataset_filter}
To prevent training the neural network on a biased dataset that contains mostly observations with no precipitation, subsetting the full dataset is necessary. Keeping observations of clear skies in the dataset might negatively affect resulting nowcasts, causing overfitting to observations containing no precipitation activity.

To filter the images, we convert CMAX reflectivity maps to rainfall rate, using the Marshall-Palmer formula from Theorem \ref{the:m-p}. Then, we compute the ratio of rainy to clear pixels in the map with the threshold of $0.05 \si{mm} / 5 \si{min}$ or $0.6 \si{mm.h^{-1}}$. Finally, if the rainfall rate map contains at least $5\%$ of rainy pixels and $11$ previous observations are available (last hour), it is added to the target observations set.

For training LUPIN, we decided to use $6$ input observations and $6$ output observations, therefore, for predicting the target observation, $11$ lead observations are needed to be kept in the dataset.

The dataset is split into train, test, and validation subsets. The last $20\%$ of target observations are selected for the test set. This set contains rainfall maps from the 2nd of September 2018 until the 30th June of 2019. The first $80\%$ of the data is chosen for training, out of which $15\%$ is split further into a validation set.

The training set is split randomly into training and validation subsets, ensuring that there are no overlapping observations in these sets (e.g. there are no target observations in the training set that are lead observations for the validation set and vice versa). The exact observation counts can be observed in Table \ref{tab:dataset_counts}.

\begin{table}[ht]
\caption{The observation counts of the full dataset, the subset selected for training according to the dataset filtering described in Section \ref{sub:dataset_filter}, and the sizes of train, test, and validation splits.
}
\label{tab:dataset_counts}
\vskip 0.15in
\begin{center}
\begin{small}
\begin{sc}
\begin{tabular}{lrrr}
\toprule
Set & \textnumero~obs. & Full \%  & Target \% \\
\midrule
Full Dataset & $355 761$ & $100.00$ & - \\
Target \& Lead Obs. & $63 007$ & $17.71$ & - \\
Target Obs. & $54 721$ & $15.38$ & $100.00$\\
Target Obs. Train. & $35 767$ & $10.05$ & $65.37$ \\
Target Obs. Valid. & $7 998$ & $2.25$ & $14.61$ \\
Test Target Obs. & $10 956$ & $3.08$ & $20.02$ \\
\bottomrule
\end{tabular}
\end{sc}
\end{small}
\end{center}
\vskip -0.1in
\end{table}

\section{Motion Field Comparison}

Before we move to the evaluation of the model as a whole, it is worth comparing only the motion fields generated by the Motion Field U-Net with the aforementioned Lukas-Kanade algorithm~\cite{lucas1981iterative} it replaces, more specifically, the pyramidal implementation available in the pySTEPS library~\cite{gmd-12-4185-2019} based on~\cite{bouguet2001pyramidal}. We will test the hypothesis that our model can produce motion fields that are better suited for the Lagrangian tranfrormatuion task.

\begin{figure*}[ht]
\centering
\subfloat[]{\includegraphics[width=0.33\textwidth]{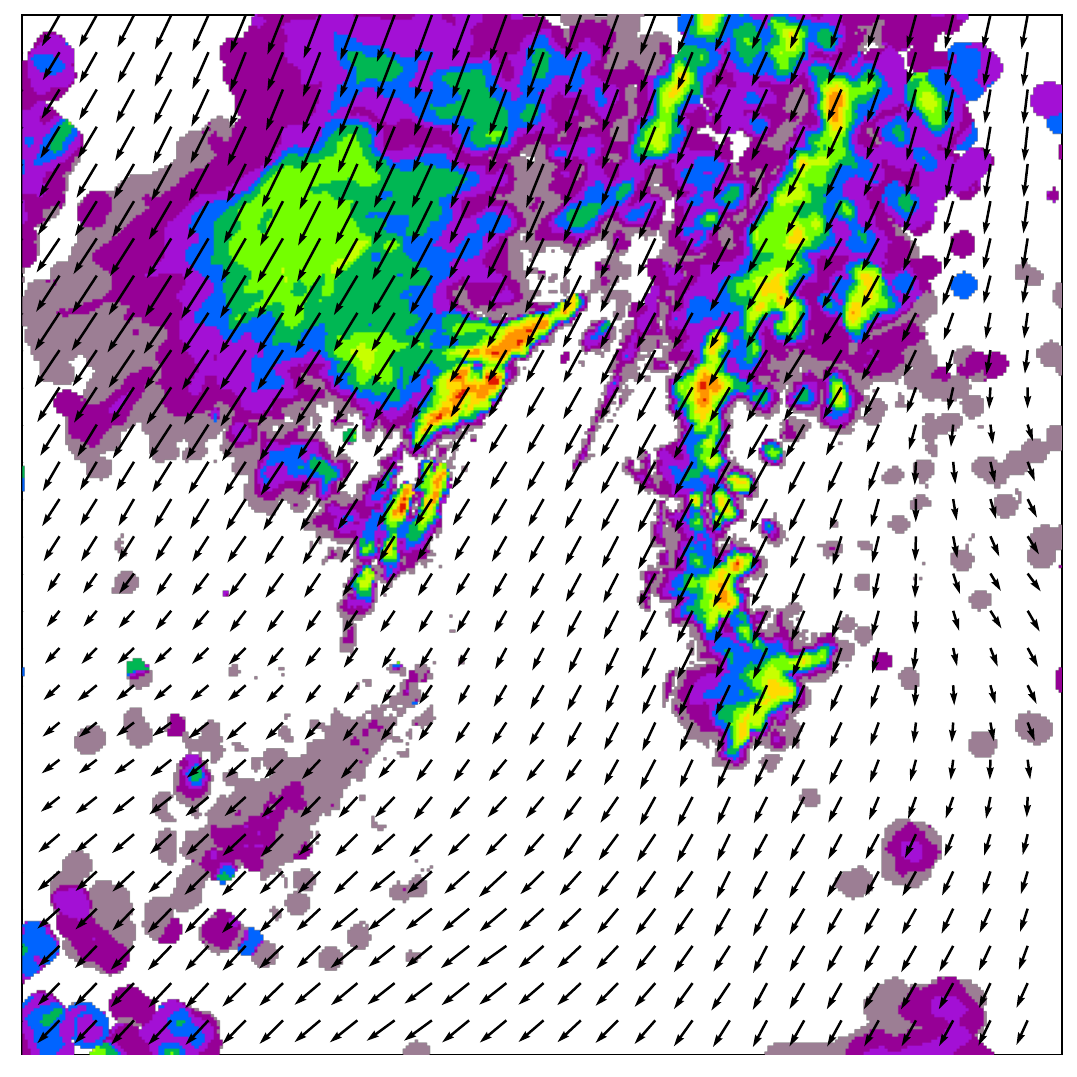}%
\label{mflk}}
\hfil
\subfloat[]{\includegraphics[width=0.33\textwidth]{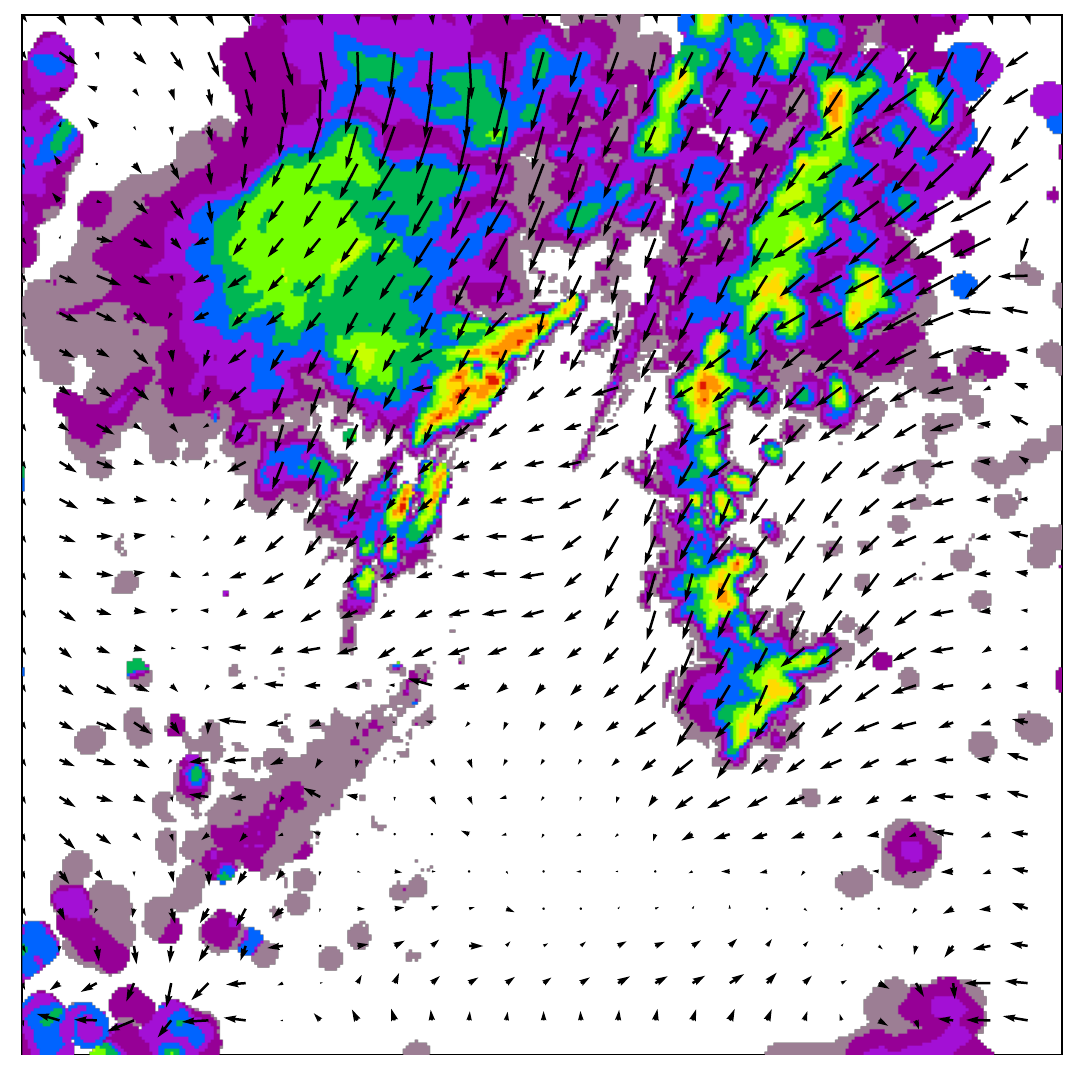}%
\label{mfnoreg}}
\hfil
\subfloat[]{\includegraphics[width=0.33\textwidth]{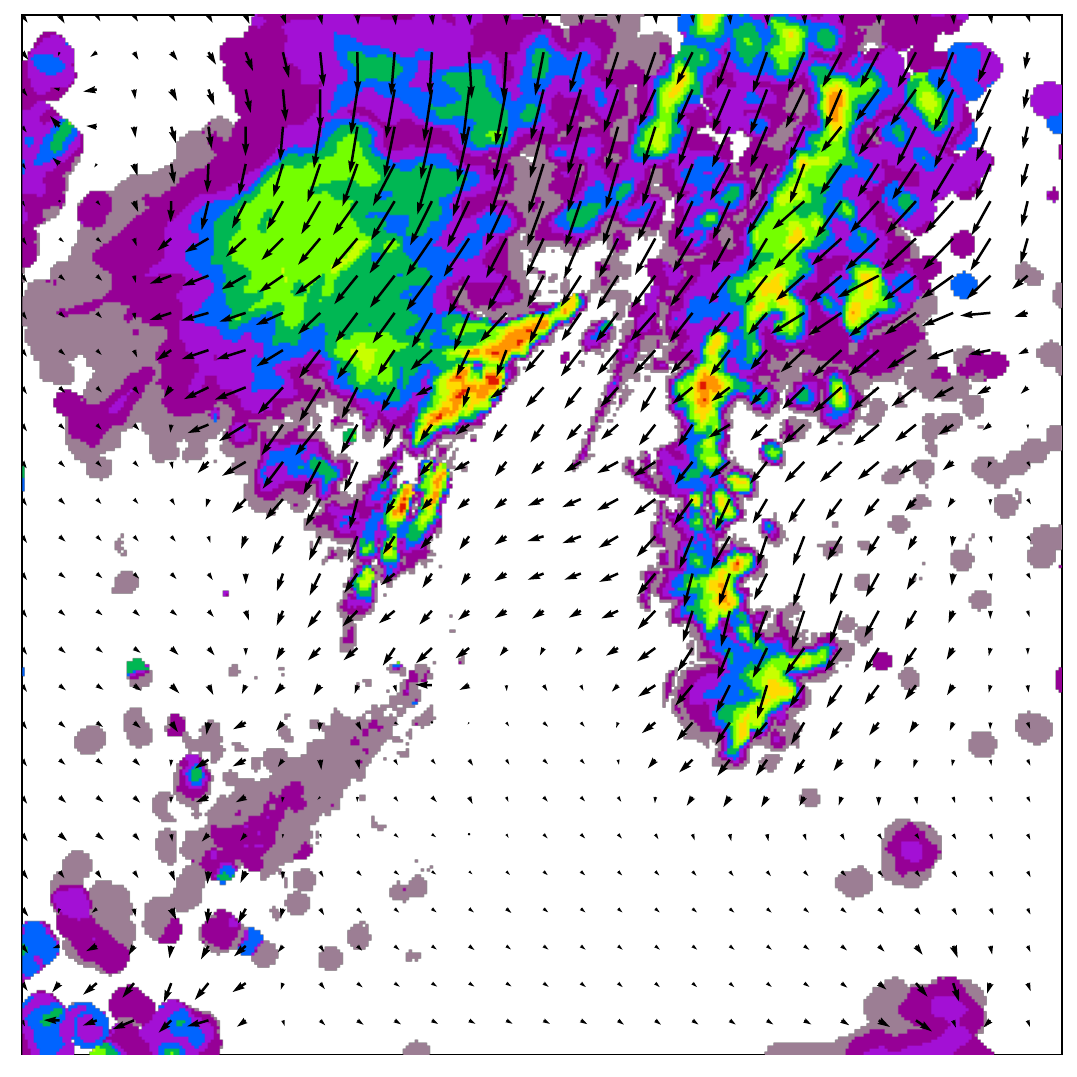}%
\label{mfwreg}}
\caption{The motion fields generated from three models for a sample extreme event from the test set. \ref{mflk}: Lukas-Kanade algorithm (the basis of L-CNN's Lagrangian transformation), \ref{mfnoreg}: MF-U-Net from LUPIN trained without the physics-informed regularization, \ref{mfwreg}: MF-U-Net from LUPIN trained with the physics-informed regularization.}
\label{fig:mfs}
\end{figure*}

We will also use this section to compare the effect of the physics-informed regularization of the motion field we apply during training. Consequently, we will compare three models -- the Lukas-Kanade algorithm, the MF-U-Net trained without the physics-informed regularization, and the one trained with it. To visually compare the motion fields from a sample extreme event from the test set, see Figure~\ref{fig:mfs} (for more information about the event, see section \ref{ssec:extreme}).

While the general direction of all the motion fields is the same, notable differences can be observed visually, especially between the motion fields of Lukas-Kanade and the MF-U-Nets. First, it's worth mentioning that Lukas-Kanade produces uniform motion fields across the entire domain (Fig. \ref{mflk}). In contrast, MF-U-Net generates only negligible motion vectors in areas with no precipitation to extrapolate (Fig. \ref{mfnoreg}, \ref{mfwreg}). However, this discrepancy does not impact the Lagrangian transformation as extrapolating these zero-valued areas is of no consequence.

On closer look, we can see more variability in the motion fields generated by MF-U-Nets. In the output of the non-regularized model (Fig. \ref{mfnoreg}), we can see divergent vectors close by in multiple areas. The regularization seems to partly eliminate this behavior as intended (Fig. \ref{mfwreg}). 

\subsection{Fitness Metrics}

To compare the models objectively, we will evaluate the motion fields generated by the models over the whole test set. However, choosing the right metrics to assess their fitness for Lagrangian transformation is not straightforward.

Ideally, the generated motion fields should be locally smooth, without any apparent sources and sinks. This is desirable not only to satisfy the continuity equation but -- in the context of the LUPIN and L-CNN models -- to preserve the precipitation field intensities during the extrapolation along the generated advection motion field since it is the role of the following advection-free U-Net to predict the temporal differences of the precipitation field in Lagrangian coordinates. Because of this, we will use the absolute value of divergence as a measure of the fitness of the motion field -- same as the divergence regularization loss from Equation \ref{eq:reg}. The lower the divergence, the better the motion field.

On the other hand, we do not want the generated motion field to be so homogeneous that it cannot accurately capture the movement of precipitation. In an extreme case, a zero-only motion field is perfectly divergence-free, but useless for performing the transformation to Lagrangian coordinates. The motion fields also need to accurately extrapolate the given input precipitation fields to the successive observations. As there is no ground truth to compare with, we will simply measure how well the last input precipitation field extrapolated by the generated motion fields matches the target observations -- measure the error of a simple Lagrangian persistence nowcast using the generated motion fields. The second metric for measuring the error will be a mean square error of the extrapolation.

\begin{figure}[ht]
    \centering
    \includegraphics[width=0.51\columnwidth]{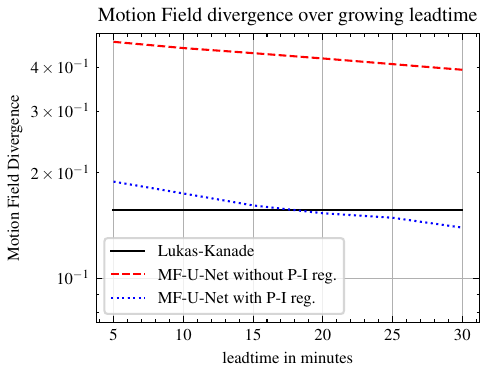}
    \includegraphics[width=0.48\columnwidth]{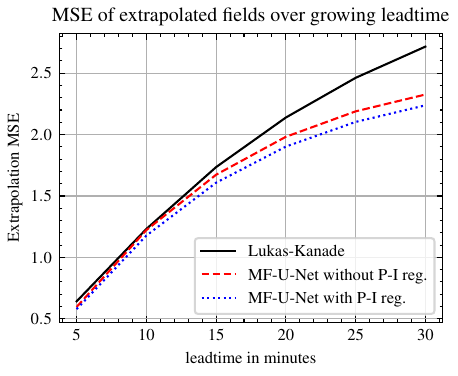}
    \caption{Average motion field divergence (left) and extrapolation MSE (right) for the whole test set, evaluated for lead times up to 30 minutes ahead. As the Lukas-Kanade produces a singular motion field that is used at every time step, the divergence is constant.}
    \label{fig:mfdivmse}
\end{figure}

The quantitative results are presented grouped by lead time to get a better idea about the models' behavior with increasing lead times. For each lead time, the last input observation is extrapolated by a set number of time steps. For Lukas-Kanade, this is straightforward, as we only consider a singular motion field for every time step. The MF-U-Net models, on the other hand, work iteratively and generate a new motion field for each time step in a rolling window fashion. See the Figure \ref{fig:mfdivmse} for the results.

From the divergence comparison, we can see that without the physics-informed regularization that penalizes highly divergent outputs, the divergence is higher, which is not surprising. However, it is good to see that the regularization pushes the model to generate motion fields on par with Lukas-Kanade in terms of divergence. Also, both MF-U-Net models exhibit slightly decreasing motion field divergence over growing leadtime.

The second graph shows a clear distinction between the models in terms of the accuracy of precipitation field extrapolation using the generated motion fields. The simple nowcasts using MF-U-Net outputs more closely match the target precipitation fields in comparison to Lukas-Kanade. In addition, the regularized MF-U-Net outperforms the unregularized one, further proving the value of the physics-informed regularization.

In conclusion, while it is hard to draw a definitive conclusion about which model generates the best motion fields for the task of Lagrangian transformation, the MF-U-Net with physics-informed regularization definitely performed well in comparison to the benchmark -- the Lukas-Kanade algorithm. With the MF-U-Net outperforming it in the extrapolation MSE metric, it could even be argued that it is better suited for the task.

\section{Nowcasting Ability Evaluation}

This section presents qualitative and quantitative results of the LUPIN training in comparison to selected reference models. As mentioned in Section \ref{sec:intro}, a visual qualitative comparison of produced nowcasts in an expert study is often more valuable than comprehensive quantitative evaluation, as there is no clear numerical metric that measures the fitness of nowcasts.

\subsection{Reference Models}

The main model to compare against is naturally L-CNN, the model LUPIN was based on. The performance of LUPIN needs to at least match the performance of the L-CNN, or, if our data-driven Lagrangian coordinate transformation is better, exceed it.

The second reference model chosen is RainNet, a data-driven U-Net-based nowcasting model. The model parameters match the U-Nets in LUPIN and L-CNN exactly. However, it is a pure machine learning model with no Lagrangian assumptions. It processes the input radar data in the usual -- Euler -- coordinates. RainNet is therefore a fitting benchmark, as any improved performance can be directly attributed to the Lagrangian approach. It also serves as a kind of ablation study of the designed Lagrangian framework.

The task of all three models is to take 6 successive precipitation fields and predict the next 6 fields. All the compared models have the same back-end structure -- an identical U-Net. Any differences observed are therefore the consequence of the different training regimes and data transformations.

\subsection{Extreme event case study} \label{ssec:extreme}

\begin{figure*}[ht]
    \centering
    \includegraphics[width=0.86\textwidth]{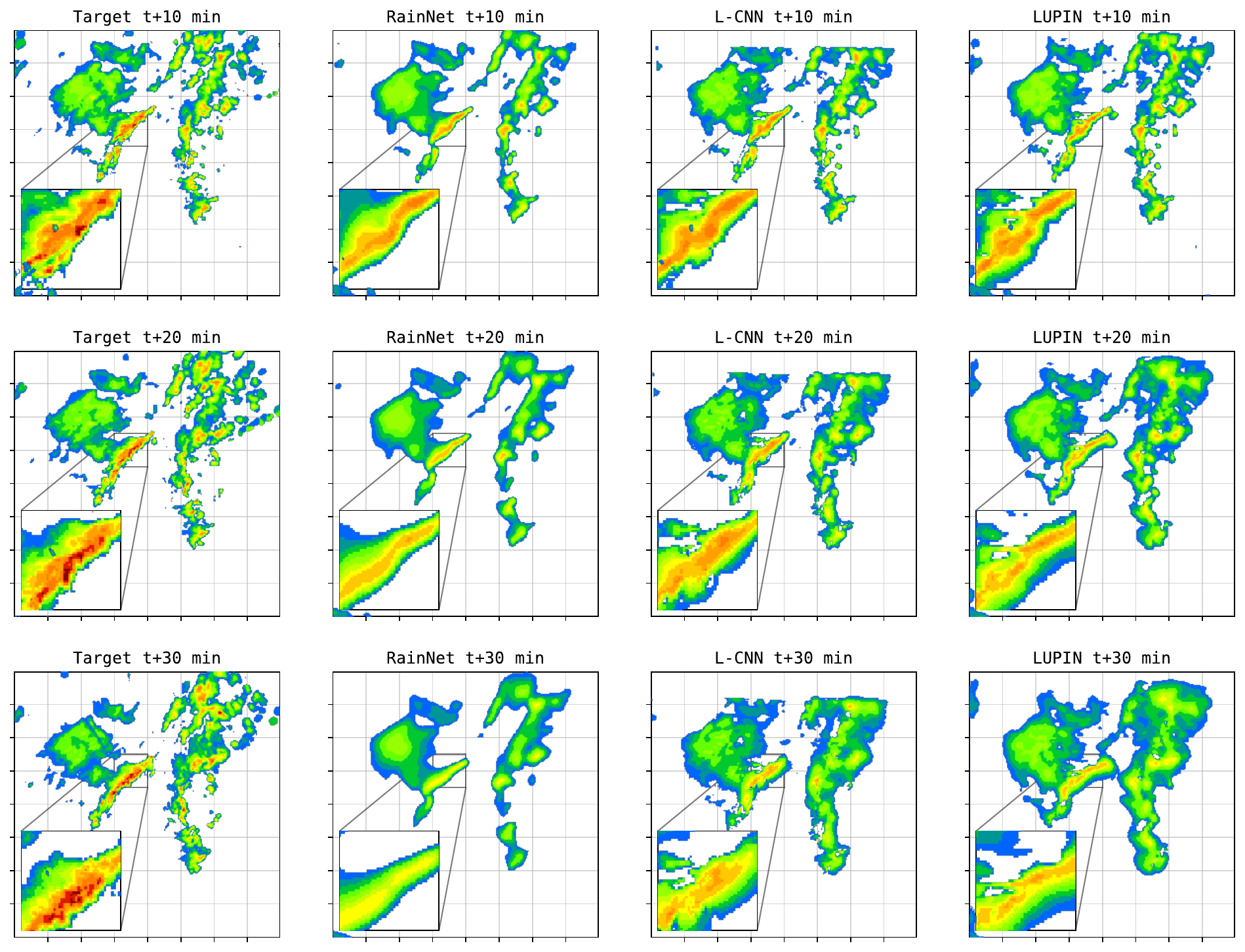}
    \includegraphics[width=0.13\textwidth]{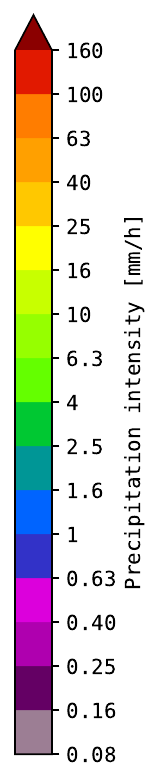}
    \caption{Extreme event example from the evening of September 2nd, 2018. The leftmost column shows radar rainfall fields of a growing convective rain cell, with the maximum rain rate region zoomed in. Precipitation below 1 mm/h was filtered out. The other columns show nowcasts of the respective target fields at three lead times: $t+10$, $t+20$ and $t+30$ minutes. All the models are able to capture the general southwestern movement of the precipitation field. However, all the models show blurring of the output over time due to uncertainty. The most prominent blurring occurs in the RainNet nowcasts, but L-CNN and LUPIN also exhibit noticeable blurring.}
    \label{fig:comp}
\end{figure*}

On September 2nd, 2018, southwest Slovakia was hit by a severe convective storm. It flooded underpasses, streets, and basements of buildings and caused public transport collapse in the capital of Slovakia~\cite{ESWD}. The convective storms are complex and hard to predict. They are the most relevant cause of systematic errors in weather and climate models~\cite{reynolds2020wgne}. This rainfall event from the test set was chosen to compare the performance of selected nowcasting models. See Figure \ref{fig:comp} for the nowcasts produced by the models.

The RainNet model exhibits very strong blurring of the output with growing lead time. In comparison, both the L-CNN and LUPIN produce less blurry outputs of similar local detail that better preserve the expected high precipitation intensity, more closely resembling the target precipitation field. The benefits of the Lagrangian approach are clearly visible.

\subsection{Quantitative Comparison}

\begin{figure}[ht]
    \centering
    \includegraphics[width=0.466\columnwidth]{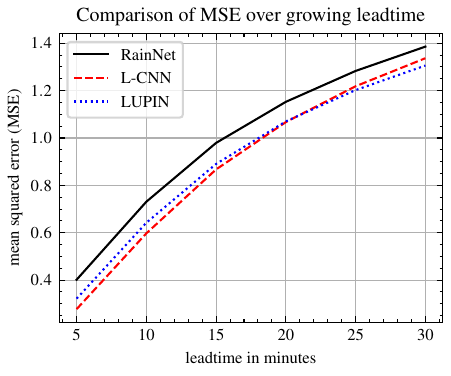}
    \includegraphics[width=0.49\columnwidth]{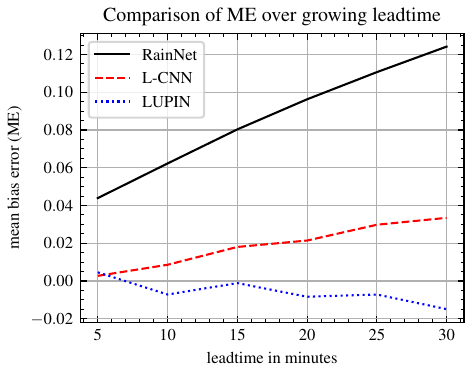}
    \caption{Plots show the test set MSE (top) and ME (bottom) of the compared models in relation to nowcast lead time.}
    \label{fig:mse and mse}
\end{figure}

To compare the models quantitatively, their performance is evaluated using multiple different metrics over the whole test set. The quantitative results are presented grouped by lead time to get a better idea about the models' behavior with increasing lead times. First, for the computed mean square error (MSE) and mean bias error (ME), see Figure~\ref{fig:mse and mse}.

The L-CNN starts best in terms of MSE, though with growing lead times LUPIN catches up and outperforms it. Both L-CNN and LUPIN exhibit relatively small bias compared to RainNet. The L-CNN model accumulates small positive bias with growing lead time and LUPIN, on the other hand, accumulates negative bias.

\begin{figure*}[!t]
\centering
\subfloat[]{\includegraphics[width=0.33\textwidth]{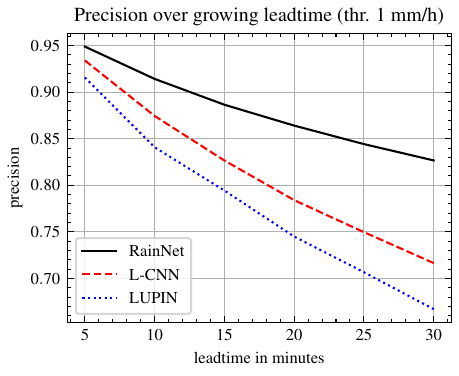}%
\label{precision1}}
\hfil
\subfloat[]{\includegraphics[width=0.33\textwidth]{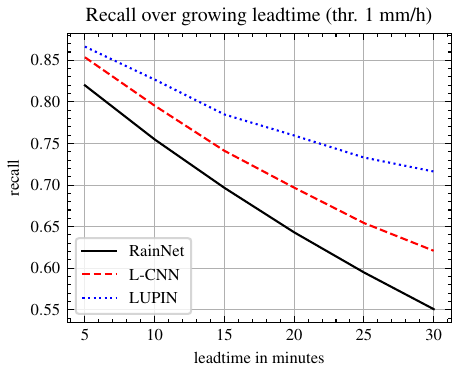}%
\label{recall1}}
\hfil
\subfloat[]{\includegraphics[width=0.33\textwidth]{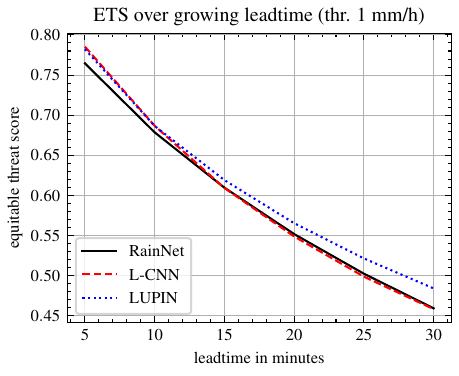}%
\label{ets1}}
\hfil
\subfloat[]{\includegraphics[width=0.33\textwidth]{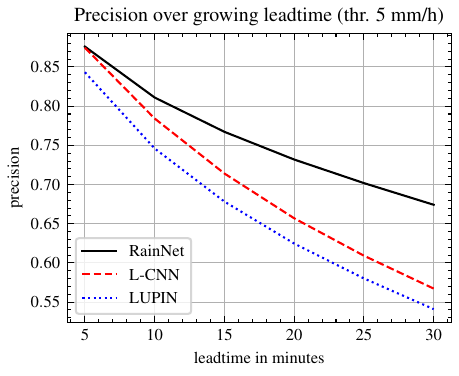}%
\label{precision5}}
\hfil
\subfloat[]{\includegraphics[width=0.33\textwidth]{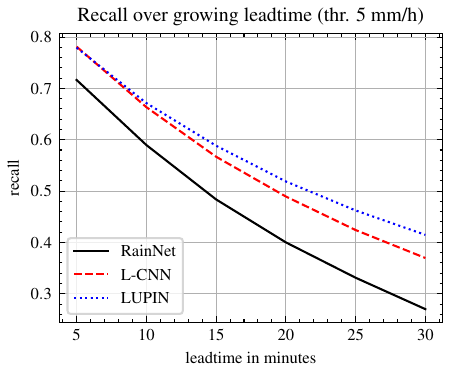}%
\label{recall5}}
\hfil
\subfloat[]{\includegraphics[width=0.33\textwidth]{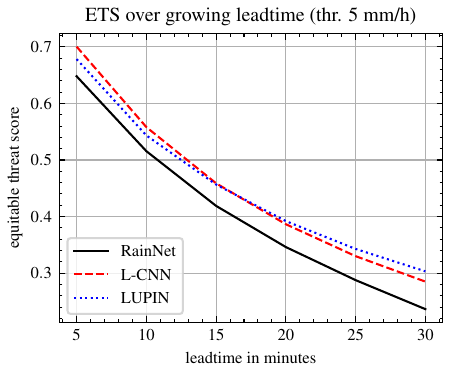}%
\label{ets5}}
\hfil
\subfloat[]{\includegraphics[width=0.33\textwidth]{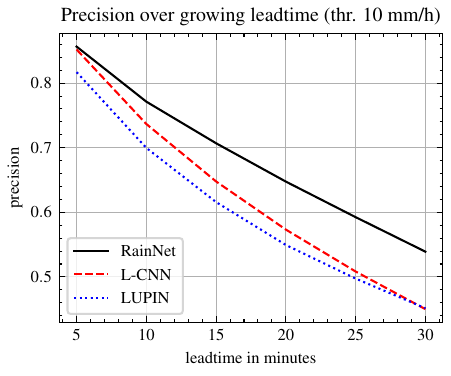}%
\label{precision10}}
\hfil
\subfloat[]{\includegraphics[width=0.33\textwidth]{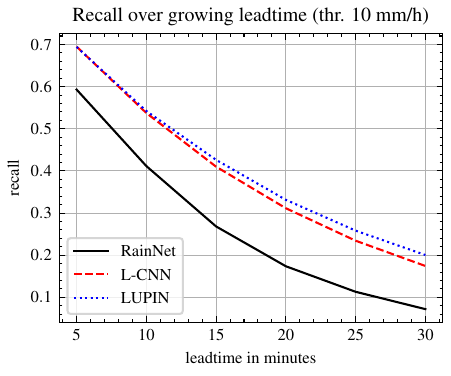}%
\label{recall10}}
\hfil
\subfloat[]{\includegraphics[width=0.33\textwidth]{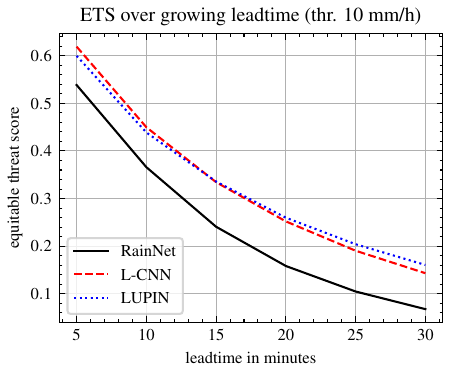}%
\label{ets10}}
\caption{Plots show test set precision (\ref{precision1}, \ref{precision5}, \ref{precision10}), recall (\ref{recall1}, \ref{recall5}, \ref{recall10}), and equitable threat score (\ref{ets1}, \ref{ets5}, \ref{ets10}) of the compared models in relation to nowcast lead time. Each row plots these three metrics for a different rain rate threshold -- 1 mm/h (\ref{precision1}, \ref{recall1}, \ref{ets1}), 5 mm/h (\ref{precision5}, \ref{recall5}, \ref{ets5}), and 10 mm/h (\ref{precision10}, \ref{recall10}, \ref{ets10}).}
\label{fig:binary}
\end{figure*}

To better evaluate the performance of the models at predicting precipitation events of various levels, we also calculate three binary classification metrics over binary rainfall fields with thresholds of 1, 5, and 10 mm/h, corresponding to light, moderate, and heavy rain respectively. We calculate precision and recall of the models along with equitable threat score, a common indicator of warning skill in weather forecasting \cite{schaefer1990critical}. See Figure~\ref{fig:binary} for comparison.

Precision -- a complement of false alarm ratio (FAR) -- marks the probability that a positive prediction (an alarm) is really a positive in the target observation. In this metric, the RainNet naturally leads due to its blurring and consequent underestimation. When it produces an alarm, there is a high chance of it being true. However, pure precision is not really of the highest priority for a warning system.

On the other hand, recall -- also called the probability of detection (POD) --  signifies the probability that a positive (an extreme event) will, in fact, trigger a positive prediction (an alarm). In the context of a warning system, this metric is of high priority to minimize the amount of missed events. Here, L-CNN and LUPIN both perform much better than RainNet. With increasing lead time, LUPIN achieves better recall compared to L-CNN, with more than 14 \% better recall at the $t+30$ minutes lead time for the 10 mm/h threshold.

Finally, the equitable threat score (ETS) -- based on the frequently used critical success index (CSI) -- considers both the false alarms and the missed events, while also taking into account hits expected by chance. Here, the RainNet also falls far behind the Lagrangian models. In this metric, the L-CNN performs better at the start, but LUPIN consistently outperforms it with increasing lead times, still ending up with 12 \% better performance at the final lead time for the 10 mm/h threshold.

\subsection{Summary}

Both the quantitative and qualitative evaluation have shown the Lagrangian approach can provide a big improvement in performance of a nowcasting model in contrast to a purely data-driven RainNet model. The L-CNN and LUPIN perform similarly, proving our framework is able to match the performance of the L-CNN while allowing much faster inference and convenient training routine without transforming the whole dataset to Lagrangian coordinates in advance. In quantitative comparison, our LUPIN model consistently outperforms the other two models in the two key warning metrics -- recall and equitable thread score -- at all rain rate thresholds. 

\section{Conclusion}

By implementing the temporal differencing applied to rainfall fields in Lagrangian coordinates introduced by the L-CNN in a fully differentiable manner, we have eliminated most of the drawbacks of the original model, while matching -- even exceeding -- its performance in key metrics.

Thanks to our implementation of the semi-Lagrangian extrapolation and training the data-driven MF-U-Net to produce physically consistent advection motion fields dynamically during runtime, we mitigate the higher computational cost at the inference of the L-CNN model. The MF-U-Net also mitigates the dependence on the quality of the motion field generated by traditional optical flow methods that can be impacted by edge effects or gaps in radar coverage in the precipitation fields.

The quality of LUPIN's transformation to Lagrangian coordinates compared to L-CNN is hard to quantify. However, we have shown that the generated advection motion fields can match the precipitation at successive observations with lower error while exhibiting similarly low divergence when trained with the physics-informed regularization loss. Combined with better overall nowcasting performance of LUPIN with increasing lead time observed in MSE, recall, and ETS metrics compared to the L-CNN, we believe this shows the motion fields learned by LUPIN are a better fit for the task of Lagrangian coordinate transformation of input data than the ones produced by the Lukas-Kanade algorithm.

The LUPIN framework, as defined in this paper, is flexible and not tied to any specific neural network architecture. With the generative models achieving great results in the precipitation nowcasting domain in recent years, our work opens the possibility of them leveraging the Lagrangian approach efficiently. We believe that including domain knowledge and physical laws in the machine learning models is necessary to build more trustworthy and reliable models of the future.

\section*{Code availability}
The source code for all the models used in the experiments, along with the training, evaluation and visualization scripts, plus animated figures are available at:
\url{https://github.com/kinit-sk/LUPIN}

\section*{Acknowledgements}
This research was partially supported by V4Grid, an Interreg Central Europe Programme project co-funded by the European Union, project No. CE0200803 and in cooperation with Softec, Ltd. and Slovak Hydrometeorological Institute.

\bibliographystyle{elsarticle-num} 
\bibliography{mybib.bib}

\end{document}